%% file: paper.tex
\documentclass[10pt,twocolumn,letterpaper]{article}

\usepackage{cvpr}              

\usepackage{graphicx}
\usepackage{amsmath}
\usepackage{amssymb}
\usepackage{booktabs}
\usepackage[table, svgnames, dvipsnames]{xcolor}
\usepackage{bbding}

\usepackage{graphicx}       
\usepackage{multirow}
\usepackage{color}
\usepackage{kotex}
\usepackage{kotex-logo}
\usepackage{algorithm}
\usepackage{algpseudocode}
\usepackage{arydshln}
\usepackage{cite}

\usepackage{caption}
\usepackage{adjustbox}
\usepackage{wrapfig}
\usepackage{lipsum}
\usepackage{makecell}

\usepackage[pagebackref,breaklinks,colorlinks]{hyperref}

\usepackage[capitalize]{cleveref}
\crefname{section}{Sec.}{Secs.}
\Crefname{section}{Section}{Sections}
\Crefname{table}{Table}{Tables}
\crefname{table}{Tab.}{Tabs.}

\def\confName{CVPR}
\def\confYear{2022}

\author{Seokju Yun\qquad Youngmin Ro\\Machine Intelligence Laboratory, University of Seoul, Korea \\ \tt\small \{wsz871, youngmin.ro\}@uos.ac.kr}

\begin{document}

\title{Dynamic Mobile-Former: Strengthening Dynamic Convolution with
Attention and Residual Connection in Kernel Space}




\maketitle

\input{0_abstract}
\input{1_intro}

\input{2_related_work}

\input{3_Method}

\input{4_result}

\input{5_conclusion}

\small
\bibliographystyle{unsrt}
\bibliography{refs}

\end{document}

%% file: 0_abstract.tex
\begin{abstract}
We introduce Dynamic Mobile-Former(DMF), maximizes the capabilities of dynamic convolution by harmonizing it with efficient operators.
Our Dynamic Mobile-Former effectively utilizes the advantages of Dynamic MobileNet (MobileNet equipped with dynamic convolution) using global information from light-weight attention.
A Transformer in Dynamic Mobile-Former only requires a few randomly initialized tokens to calculate global features, making it computationally efficient.
And a bridge between Dynamic MobileNet and Transformer allows for bidirectional integration of local and global features.
We also simplify the optimization process of vanilla dynamic convolution by splitting the convolution kernel into an input-agnostic kernel and an input-dependent kernel. 
This allows for optimization in a wider kernel space, resulting in enhanced capacity. 
By integrating lightweight attention and enhanced dynamic convolution, our Dynamic Mobile-Former achieves not only high efficiency, but also strong performance.
We benchmark the Dynamic Mobile-Former on a series of vision tasks, and showcase that it achieves impressive performance on image classification, COCO detection, and instanace segmentation.
For example, our DMF hits the top-1 accuracy of 79.4\% on ImageNet-1K, much higher than PVT-Tiny by 4.3\% with only $1/4$ FLOPs.
Additionally, our proposed DMF-S model performed well on challenging vision datasets such as COCO, achieving a 39.0\% mAP, which is 1\% higher than that of the Mobile-Former 508M model, despite using 3 GFLOPs less computations. 
Code and models are available at  \url{https://github.com/ysj9909/DMF}
\end{abstract}





%% file: 1_intro.tex
\section{Introduction}\label{intro}
Recently, vision transformer(ViT)~\cite{vit,deit} demonstrates the advantage of global processing and excellent capabilities with scalability over convolutional neural networks (CNNs).
However, since ViT has a time complexity that scales quadratically with the sequence length, its computational efficiency decreases when working with longer sequences, making it more challenging to train and deploy ViT models in scenarios with limited computational resources.
If we reduce the computational cost to under 500M FLOPs, MobileNet~\cite{mobilenet1, mobilenetv2, mobilenetv3} and other similar-sized lightweight CNNs~\cite{shufflenet, shufflenetv2, efficientnet} still outperform ViT variants in terms of efficiency, thanks to their ability to perform local processing using techniques such as depth-wise separable convolution~\cite{mobilenetv2} and group convolution with channel shuffle~\cite{shufflenet, shufflenetv2}.


Along with great advances in efficient CNN architecture design, dynamic convolution~\cite{yang2019condconv, chen2020dynamic} has recently gained popularity for the implementation of lightweight networks due to its ability to achieve significant performance gains with negligible computational cost. 
This has led to its adoption for multiple vision tasks~\cite{chen2020dynamic, chen2021dynamicregion, ma2020weightnet, tian2020conditionalseg, he2021dyco3d, liu2022dynamicprototype}.
Moreover, as many computer vision applications are not constrained by the number of parameters, but instead have strict latency requirements for inference, such as real-time object detection and tracking for autonomous driving, the potential of using dynamic convolution in these scenarios is increasing.

The basic idea of dynamic convolution is to dynamically aggregate multiple convolution kernels into a convolution weight matrix, using an input-dependent attention mechanism.
\small
\begin{equation}\label{eq:vanilla dyconv}
\begin{aligned}
        \mathbf{W}_{dynamic}(x) = \sum_{k=1}^K \pi_k(x)\mathbf{W}_{static}^k \quad s.t. \quad 0 \le \pi_k(x) \le 1, 
\end{aligned}
\end{equation}
\normalsize

\noindent where K static convolution kernels {$\mathbf{W}_{static}^k$} are weighted summed using attention scores { $\pi_k(x)$}.

\noindent Dynamic convolution has two limitations: 1) The input to a kernel attention module may have limited representational power, which can restrict the ability of the attention mechanism to calculate  relevant filters effectively. 2) a joint optimization of attention scores and static kernels can be challenging.
In this work, we propose our dynamic residual group convolution that can efficiently compute input-specific local features while addresses the limitations mentioned above.
\twocolumn[{%
\renewcommand\twocolumn[1][]{#1}%
\begin{center}
    \centering
    \vspace{-15mm}
    \captionsetup{type=figure}
    \includegraphics[width=1.0\linewidth]{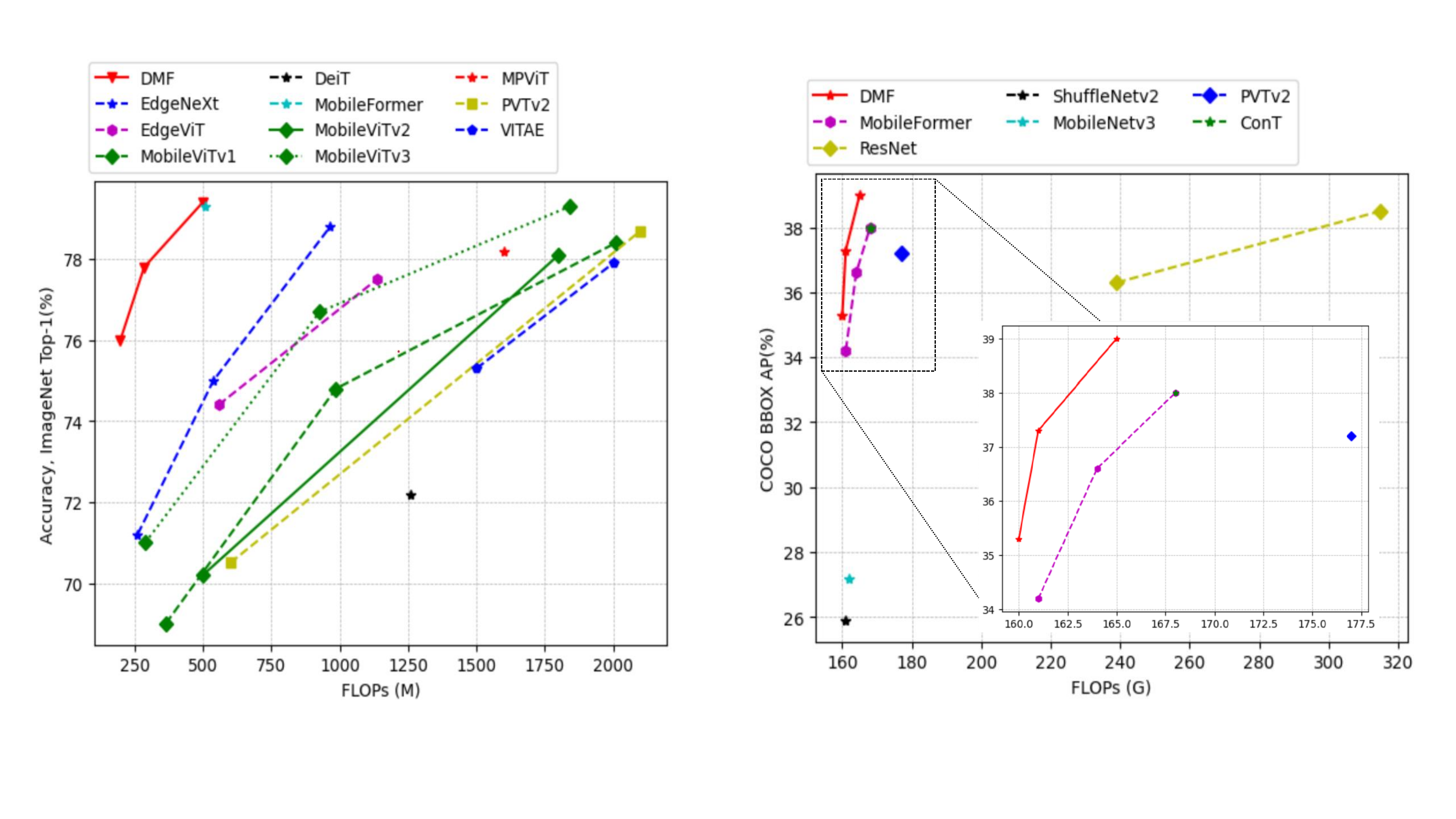}
    \vspace{-18mm}
    \captionof{figure}{\textbf{Performance comparison between DMF and other methods.} Left: Top-1 accuracy on ImageNet-1K~\cite{deng2009imagenet}. Right: Object detection results on COCO val2017~\cite{lin2014microsoftcoco} of various backbones using RetinaNet~\cite{lin2017focal} framework, all numbers are for single-scale training, 12 epochs (1$\times$) training schedule. Best viewed in color.}
    \label{fig:plot}
\end{center}%
}]
\noindent~\cite{chen2020dynamic, yang2019condconv} uses the global average pooled features as inputs to the kernel attention module to generate attention scores. However, spatially squeezed features have limited representational power, because all features at each location are merged with the same weight. To solve this limitation, our Dynamic Mobile-Former utilizes global salient features, which are calculated using light-weight attention, as input to the kernel attention module. This allows appropriate kernels to be selected for each input.

\noindent And ~\cite{yang2019condconv} proposed the use of a sigmoid layer to generate attention scores, leading to a significantly large space for the convolution kernel that makes the learning of attention scores difficult.
~\cite{chen2020dynamic} replaced the sigmoid layer with a softmax function to compress the kernel space, thereby facilitating learning of the attention module.
However, the model capacity is reduced as the kernel space is limited.
To solve this problem, we separate the existing convolution kernel into an input-agnostic kernel and an input-dependent kernel to ease learning of the kernel attention module. In addition to our method, we use sigmoid activation to achieve better performance than ~\cite{chen2020dynamic, yang2019condconv}.
Its effectiveness has been verified by experimental results in section~\ref{ablations}.
We also propose a more efficient module by combining the above techniques with group convolution(see section~\ref{sec: dy-mobile}).

\noindent Our DMF model achieves superior performance in terms of both accuracy and FLOPs, as demonstrated through extensive experiments on ImageNet and other downstream tasks.
Our method shows the favorable tradeoff between accuracy and FLOPs (see Fig. \ref{fig:plot}).
For example, DMF-S achieves 79.4\% top-1 accuracy on ImageNet, which is higher than that of the MobileFormer-508M~\cite{mobileformer} (current state-of-the-art) with less computations.
DMF-S also hits 0.5\% higher performance in COCO detection compared to ResNet101~\cite{he2016resnet}, while utilizing almost half the computations required by ResNet101 (315G $\rightarrow$ 165G).



%% file: 2_related_work.tex
\section{Related Work}
\noindent\textbf{Efficient CNNs: } MobileNet~\cite{mobilenet1, mobilenetv2, mobilenetv3} uses depthwise separable convolution to decompose a standard k × k convolution into a depthwise convolution and a pointwise convolution.
ShuffleNet~\cite{shufflenet, shufflenetv2} combines group convolution and channel shuffle to improve the efficiency of the network.
EfficientNet~\cite{efficientnet, efficientdet} proposes a compound scaling method that scales up the depth, width, and resolution of the network in a principled manner.
Other efficient operators include cheap linear transformations in GhostNet~\cite{ghostnet}, mixing up multiple kernel sizes~\cite{mixconv}, and using additions to trade multiplications in AdderNet~\cite{addernet}.
Our Dynamic Mobile-Former effectively combines the efficient operators such as depthwise separable convolution, group convolution presented mentioned above.

\vspace{1.5 mm}
\noindent\textbf{Vision Transformers and efficient variants: } The pioneering work ViT~\cite{vit} directly applied the transformer~\cite{attention} to classification with image patches as input.
Since then, there have been numerous attempts~\cite{conditional, tit, brain, liu2021swin, deit, wang2021pyramid, wu2021cvt, detr, deformabledetr, 3detr, xie2021segformer} to apply transformer in computer vision tasks.
Furthermore, research aimed at harmonizing strengths of ViT and CNN models~\cite{conditional, wu2021cvt, alternet, inceptionformer, guo2022cmt, xiao2021early, dai2021coatnet, xu2021co, xu2021vitae} has gained popularity in the computer vision community.
~\cite{conditional, wu2021cvt, guo2022cmt} leverage a convolutional projection into a vanilla attention~\cite{attention, vit} module.
There are studies that utilize both attention and convolutional blocks either sequentially~\cite{alternet, dai2021coatnet} or in parallel~\cite{xu2021co, xu2021vitae, inceptionformer, mobileformer}.
\twocolumn[{%
\renewcommand\twocolumn[1][]{#1}%
\begin{center}
    \centering
    \vspace{-3mm}
    \captionsetup{type=figure}
    \includegraphics[width=1.0\linewidth]{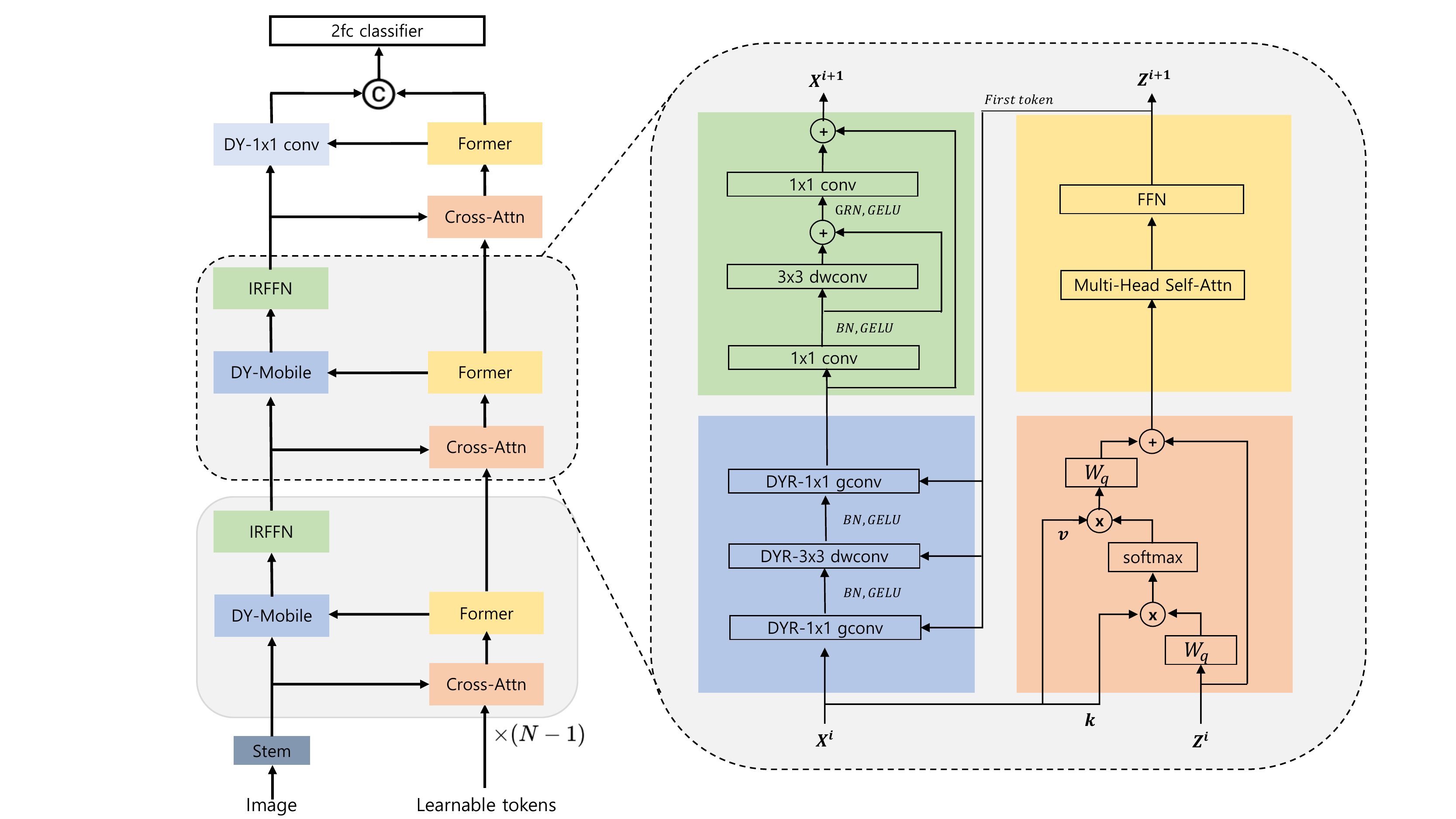}
    \captionof{figure}{\textbf{The overall architecture of Dynamic Mobile-Former(DMF) and details of DMF block}. Following~\cite{mobileformer}, Dynamic Mobile-Former adopts parallel design for processing both local and global features. Each layer includes a Dynamic Mobile block, a cross attention block, a Former block,and a Inverted Residual FFN. Also DYR-gconv and DYR-dwconv in Dynamic Mobile block means dynamic residual group convolution, dynamic residual depth-wise convolution respectively. Best viewed in color.}
    \label{fig:Model}
\end{center}%
}]
Different from the above ViT variants, there is another line of works~\cite{mobileformer, jaegle2021perceiver} that use cross-attention with very few learnable tokens.
Our Dynamic Mobile-Former also uses cross-attention with very few learnable tokens for calculating global sailent features. 
Our Dynamic Mobile-Former endows Dynamic Convolution~\cite{yang2019condconv, chen2020dynamic} to convey global information to local features, in contrast to ~\cite{mobileformer} which utilizes cross-attention.

\vspace{1.5 mm}
\noindent\textbf{Dynamic Neural Networks: } Dynamic networks increase their representation power by adjusting their parameters or activation functions~\cite{chen2020dynamicrelu} based on the input. 
~\cite{hu2018squeeze, woo2018cbam} recalibrates channel informeation by squeezing global context.
Dynamic convolution~\cite{yang2019condconv, chen2020dynamic} aggregates multiple kernels based on input dependet attention.
DCD~\cite{decomposition} proposes a dynamic convolution decompsition method which can get more compact yet competitive models to handle limitations of the dynamic convolution.
ODConv~\cite{li2022omni} introduces a dynamic convolution approach that is not only computationally efficient, but also parametrically efficient.
Instead, in this paper we also aim to address the limitations of dynamic convolution in different manner, see the Introduction and Method sections for details.

%% file: 3_Method.tex
\section{Dynamic Mobile-Former}

\subsection{Overall Architecture}
As shown in Fig.~\ref{fig:Model}, Dynamic Mobile-Former(DMF) combines the MobileNet~\cite{mobilenetv2} and Transformer~\cite{vit} using dynamic convolution and lightweight cross attention.
DY-Mobile (refers to Dynamic Mobile block) extracts local features input-dependently from an input image, while Former and Cross-Attn (refer to Transformer block and cross attention layer, respectively) extract global features using learnable tokens.
Unlike ViT, which projects local image patches linearly, Former and Cross-Attn use significantly fewer parameters (e.g. 6 or fewer tokens) resulting in reduced computational cost.
As mentioned in ~\cite{alternet}, ViTs and CNNs are complementary, and it is important to combine features of attention and convolution appropriately.
To achieve this goal, different approaches have been proposed such as adding attention layers at the end of each stage as in ~\cite{alternet}, configuring two layers in parallel and element-wise adding their outputs as in ~\cite{xu2021vitae}, and utilizing cross-attention layers as in ~\cite{mobileformer}.
Following ~\cite{mobileformer}, our DMF uses cross attention layer to fuse local features to global tokens. 
However, to fuse global tokens into local features, we input the global token into the kernel attention module in dynamic convolution~\cite{chen2020dynamic, yang2019condconv}.

To improve model efficiency, we use group convolutions on 1x1 layers, which reduces computational costs by ensuring each convolution operates on its corresponding input channel group. However, when multiple group convolutions are stacked together, a side effect can occur where the outputs from a particular channel are derived from only a small fraction of input channels.
To enable feature communication  between different groups of channels, we apply Inverted Residual Feed-Forward Network(IRFFN)~\cite{guo2022cmt} after the DY-Mobile.
Fully-connected layers in IRFFN can solve the above problem by connecting all channels of groups that are separated from each other.
The DMF block is able to capture local dependencies by generating input-specific kernel using global information.

\subsection{DMF Block}
The proposed DMF block consists of a Dynamic Mobile block(DY-Mobile), a Former block with cross attention layer, and an Inverted Residual Feed-Forward Network(IRFFN), as illustrated in Fig.~\ref{fig:Model} right.
DMF block has two inputs. 
1) Local feature map $\mathbf{X}^i \in \mathbb{R}^{N \times C}$ with $C$ channels and sequence length $N = H \times W$ (the resolution of the imput of current block), and 2) global tokens $\textbf{Z}^i \in \mathbb{R}^{M \times d}$, where $M$ and $d$ are the number and dimension of tokens, respectively. Note that $M$ and $d$ are consistent across all blocks. 
DMF block outputs the updated local feature map $\textbf{X}^{i+1}$ and global tokens $\textbf{Z}^{i+1}$, which are used as input for the subsequent block.
We will describe these Four parts in the following.

\subsection{DY-Mobile}\label{sec: dy-mobile}
DY-Mobile is built based on the inverted bottleneck in ~\cite{mobilenetv2} with three modifications.
Firstly, we substitute all vanilla convolutions with our dynamic residual convolutions.
The dynamic residual convolution kernel consists of two parts: an input-agnostic kernel and a dynamic kernel calculated as Eq.\ref{eq:vanilla dyconv}.
specifically, adding the input-agnostic kernel to vanilla dynamic kernel leads to

\small
\begin{equation}\label{eq:our dyconv}
\begin{aligned}
    \mathbf{W}_{dy-res}(x) = \mathbf{W}_{dynamic}(x) + \mathbf{W}_{input-agnostic} \quad \quad \quad \quad \\
    where \quad \mathbf{W}_{dynamic}(x) = \sum_{k=1}^K \pi_k(x)\mathbf{W}_{static}^k , \quad 0 \le \pi_k(x) \le 1,
\end{aligned} 
\end{equation}
\normalsize

\begin{figure}
    \centering
    \vspace{-3mm}
    \includegraphics[width=1.0\linewidth]{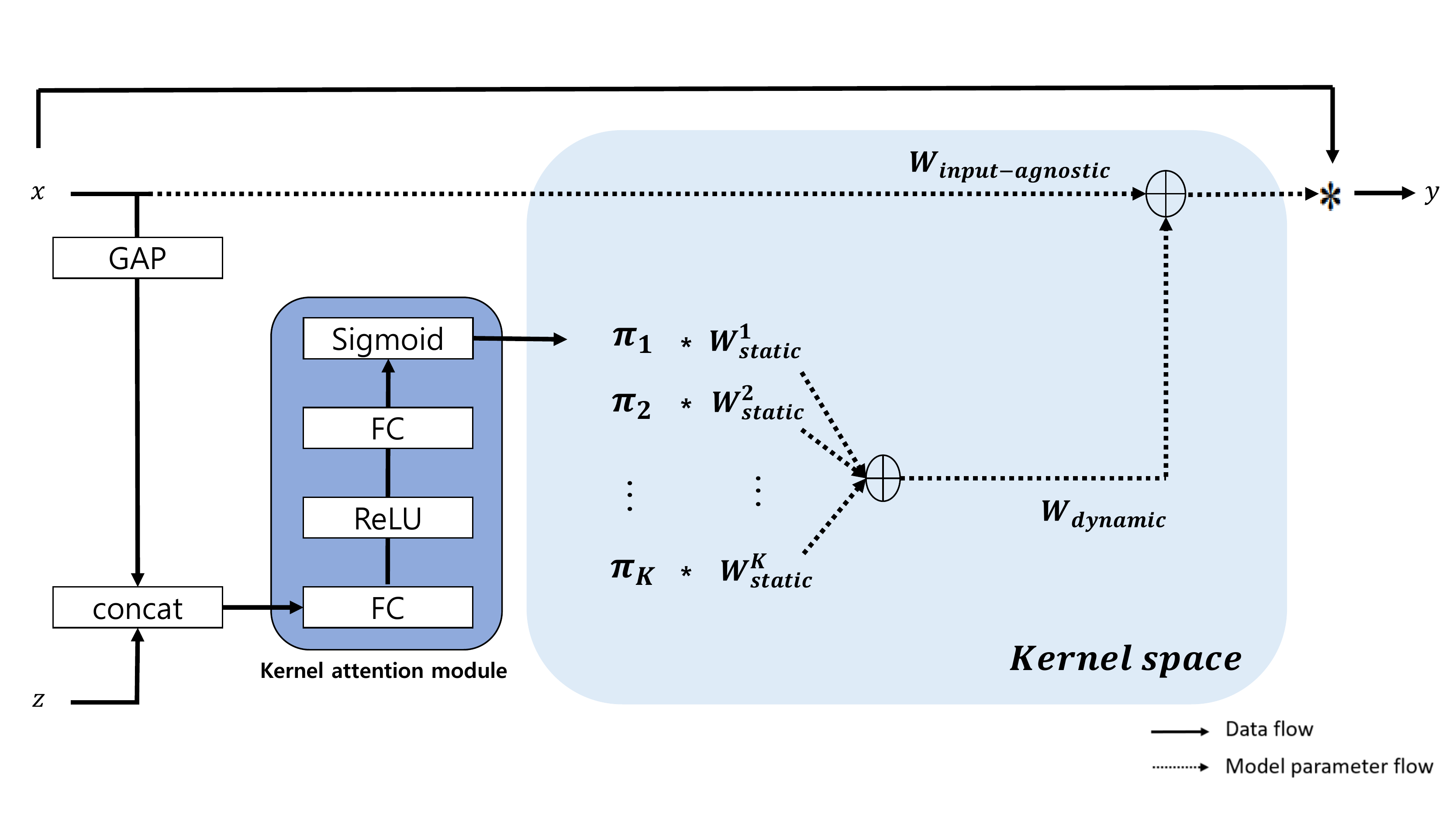}
    \caption{\textbf{A Dynamic Residual Convolution layer.} Firstly, the global average pooled features of the input and the first global token are concatenated. These concatenated features are then passed through a kernel attention module to generate attention scores ($\pi$) and obtain dynamic convolution weights using Eq.~\ref{eq:vanilla dyconv}. Finally, a dynamic residual kernel is obtained by adding the obtained dynamic kernel and the input-agnostic kernel as described in Eq.~\ref{eq:our dyconv}. And $W_{static}$ are zero initialized according to the concept of residual connection, but $W_{input-agnostic}$ is initialized randomly.}
    \label{fig:2}
\end{figure}

\noindent Where $\mathbf{W}_{static}$ and $\mathbf{W}_{input-agnostic}$ are static but  $\mathbf{W}_{static}$ is initialized to zero while $\mathbf{W}_{input-agnostic}$ is initialized randomly.
Fig.~\ref{fig:2} provides a schematic visualization of dynamic residual convolution layer.
specifically, to compute attention scores, we use sigmoid activation function. 
And following ~\cite{chen2020dynamic}, we adopt a temperature annealing strategy in the early training process to suppress the near zero output of the sigmoid function.
By doing so, all convolution kernels are optimized simultaneously in early training epochs.
Secondly, we use group convolution for point-wise convolution. 
Since a high expansion ratio(3 - 6) is used in the block~\cite{mobilenetv2}, efficiency can be greatly increased by using group convolution.
Lastly, we replace ReLU with GELU~\cite{hendrycks2016gelu}
as the activation function, following~\cite{liu2022convnext}.
Note that the kernel size of depth-wise convolution is 3$\times$3 for all layers.

\noindent DY-Mobile consumes computations of $O(NC^2)$. Along with IRFFN (specified in sec ~\ref{sec:irffn}), it accounts for the majority of the computational complexity.

\input{Table/table_cvpr/gan_test}

\subsection{Former with Cross Attention}
The light-weight cross attention from local feature map $\textbf{X}$ to global tokens $\textbf{Z}$ is computed as:

\begin{gather}
    \mathrm{CrossAttn} = \mathrm{Concat}(\mathrm{head}_1,...,\mathrm{head}_N)W^O, \\
    \mathrm{head}_i = \mathrm{Attention}(\textbf{Z}_iW^Q_i, \textbf{X}_i, \textbf{X}_i), \\
    \mathrm{Attention}(\textbf{q}, \textbf{k}, \textbf{v}) = \mathrm{Softmax}(\textbf{q}\textbf{k}^\mathsf{T} / \sqrt{d_{head}})\textbf{v},
\end{gather}
where $\mathrm{Concat}(·)$ is the concatenation operation.

\noindent$W^Q_i\in\mathbb{R}^{d \times d_{head}}$ and $W^O\in\mathbb{R}^{d \times d}$ are linear projection weights. $N$ is the head number of the attention layer. 
Therefore, the dimension of each head $d_{head}$ is equal to $\frac{d}{N}$.
To reduce computational requirements from high resolution feature map $\textbf{X}$, we eliminate the projections ($W^K$, $W^V$).
Through these formulas, the tokens $Z$ are able to learn global priors.  
The subsequent Former sub-block is a standard Transformer~\cite{attention} block including a multi-head self attention and a feed-forward network(FFN).
An expansion ratio of 2 is utilized for FFN.
The global features, obtained as described above, is fed into the dynamic residual convolutions.
When processing an input feature map of size $N \times C$ along with $M$ global tokens of dimension $d$, Former model with cross-attention has a computational complexity of $O(M^2d + Md^2 + MNC + MdC).$

\subsection{IRFFN}\label{sec:irffn}
The IRFFN sub-block differs slightly from the inverted residual FFN in ~\cite{guo2022cmt} by replacing Batch Normalization~\cite{ioffe2015batch} after shortcut connection with Global Response Normalization(GRN)~\cite{woo2023convnextv2}.
specifically, the first layer expands the dimension by a factor of a certain number, and the second layer reduces the dimension by the same ratio.
Between these two layers, the depth-wise convolution with shortcut connection is used:
\begin{gather}
    \mathrm{IRFFN}(\mathbf{X}) = \mathrm{1x1Conv}(\mathrm{SC}(\mathrm{1x1Conv}(\mathbf{X}))), \\
   \mathrm{SC}(\mathbf{X}) = \mathrm{3x3DWConv}(\mathbf{X}) + \mathbf{X},
\end{gather}
where the activation(GELU~\cite{hendrycks2016gelu}), GRN is omitted. 
We also include the batch normalization~\cite{ioffe2015batch} after two 1x1 convolutions.
This alteration facilitates communication between features that were calculated independently in each channel group of previous DY-Mobile. By using GRN~\cite{woo2023convnextv2}, various features can be calculated in different groups in subsequent DY-Mobile.
This sub-block has the same computational costs as DY-Moible ($O(NC^2)$).

\subsection{Model Specification}
table~\ref{tab: architectures} shows the detailed architectures at different computational complexities (i.e. 499M - 198M FLOPs). 
Dynamic Moible-Former(DMF) has three models with different configurations based on the number of groups in DY-Mobile.
While they share similar model designs in terms of having the same number of channels per group, they differ in the number of groups and expansion ratio in IRFFN.
As an illustration, DMF with 499M FLOPs for image size 224$\times$224, which stacks 11 DMF blocks.
All blocks contain six global tokens with a dimension of 192, following ~\cite{mobileformer}.
It begins with a 3$\times$3 convolution as stem and a lite bottleneck block~\cite{li2020micronet} in stage 1.
The bottleneck block expands and then squeezes the number of channels by stacking a 3x3 depth-wise and point-wise convolution.
For downsampling across multiple blocks, a 3$\times$3 depth-wise convolution with stride of 2 is used.
To compute head input, we concatenate global average pooled local features with first global token. These features are then passed through two fully-connected layers with hard-swish~\cite{mobilenetv3} in between.

DMF generates four hierarchical feature maps with varying resolutions, similar to ~\cite{wang2021pyramid, guo2022cmt, he2016resnet, mobilenetv2, efficientnet}.
These feature maps have strides of 4, 8, 16, and 32 with respect to the input image, enabling DMF to obtain multi-scale representations.
This makes DMF well-suited for downstream tasks such as object detection and instance segmentation, as it can capture both fine and coarse details of an object at different scales.

\input{Table/imagenet_results}

%% file: Table/table_cvpr/gan_test.tex
\begin{table*}[h!]
\centering
\resizebox{1.00\linewidth}{!}{%
\begin{tabular}{c|cllcc|cllcc|cllcc}
\Xhline{3\arrayrulewidth}
\multirow{2}{*}{stage (output size)}   & \multicolumn{5}{c|}{\textbf{DMF - S}}                                                                    & \multicolumn{5}{c|}{\textbf{DMF - XS}}                                                                 & \multicolumn{5}{c}{\textbf{DMF - XXS}}                                                                \\ \cline{2-16} 
                                       & \multicolumn{3}{c}{operator}           & \multicolumn{1}{l}{\#exp} & \multicolumn{1}{l|}{\#out} & \multicolumn{3}{c}{operator}         & \multicolumn{1}{l}{\#exp} & \multicolumn{1}{l|}{\#out} & \multicolumn{3}{c}{operator}         & \multicolumn{1}{l}{\#exp} & \multicolumn{1}{l}{\#out} \\ \hline
\multicolumn{1}{l|}{Tokens (\# heads)} & \multicolumn{5}{c|}{6 × 192 (8)}                                                                & \multicolumn{5}{c|}{6 × 192 (8)}                                                              & \multicolumn{5}{c}{6 × 192 (8)}                                                              \\ \hline
Stem (112 × 112)                       & \multicolumn{3}{c}{vanilla 3×3conv}    & -                         & 24                         & \multicolumn{3}{c}{vanilla 3×3conv}  & -                         & 16                         & \multicolumn{3}{c}{vanilla 3×3conv}  & -                         & 12                        \\ \hline
1 (112 × 112)                          & \multicolumn{3}{c}{bneck-lite}         & 48                        & 24                         & \multicolumn{3}{c}{bneck-lite}       & 32                        & 18                         & \multicolumn{3}{c}{bneck-lite}       & 24                        & 12                        \\ \hline
Downsampling                           & \multicolumn{3}{c}{3×3 dwconv}         & -                         & 24                         & \multicolumn{3}{c}{dwconv 3×3}       & -                         & 18                         & \multicolumn{3}{c}{3×3 dwconv}       & -                         & 12                        \\ \hline
2 (56 × 56)                            & \multicolumn{3}{c}{DMF block(2,4,2.7)} & 144                       & 40                         & \multicolumn{3}{c}{DMF block(2,3,2)} & 108                       & 30                         & \multicolumn{3}{c}{DMF block(2,2,3)} & 72                        & 20                        \\
2 (56 × 56)                            & \multicolumn{3}{c}{DMF block(2,4,2.7)} & 120                       & 40                         & \multicolumn{3}{c}{DMF block(2,3,2)} & 90                        & 30                         & \multicolumn{3}{c}{DMF block(2,2,3)} & 60                        & 20                        \\ \hline
Downsampling                           & \multicolumn{3}{c}{3×3 dwconv}         & -                         & 40                         & \multicolumn{3}{c}{3×3 dwconv}       & -                         & 30                         & \multicolumn{3}{c}{3×3 dwconv}       & -                         & 20                        \\ \hline
3 (28 × 28)                            & \multicolumn{3}{c}{DMF block(2,4,2.7)} & 240                       & 72                         & \multicolumn{3}{c}{DMF block(2,3,2)} & 180                       & 54                         & \multicolumn{3}{c}{DMF block(2,2,3)} & 120                       & 36                        \\
3 (28 × 28)                            & \multicolumn{3}{c}{DMF block(2,4,2.7)} & 216                       & 72                         & \multicolumn{3}{c}{DMF block(2,3,2)} & 162                       & 54                         & \multicolumn{3}{c}{DMF block(2,2,3)} & 108                       & 36                        \\ \hline
Downsampling                           & \multicolumn{3}{c}{3×3 dwconv}         & -                         & 72                         & \multicolumn{3}{c}{3×3 dwconv}       & -                         & 54                         & \multicolumn{3}{c}{3×3 dwconv}       & -                         & 36                        \\ \hline
4 (14 × 14)                            & \multicolumn{3}{c}{DMF block(2,4,2.7)} & 432                       & 128                        & \multicolumn{3}{c}{DMF block(2,3,2)} & 324                       & 96                         & \multicolumn{3}{c}{DMF block(2,2,3)} & 216                       & 64                        \\
4 (14 × 14)                            & \multicolumn{3}{c}{DMF block(4,4,2.7)} & 512                       & 128                        & \multicolumn{3}{c}{DMF block(3,3,2)} & 384                       & 96                         & \multicolumn{3}{c}{DMF block(2,2,3)} & 256                       & 64                        \\
4 (14 × 14)                            & \multicolumn{3}{c}{DMF block(4,4,2.7)} & 768                       & 176                        & \multicolumn{3}{c}{DMF block(3,3,2)} & 576                       & 132                        & \multicolumn{3}{c}{DMF block(2,2,3)} & 384                       & 88                        \\
4 (14 × 14)                            & \multicolumn{3}{c}{DMF block(4,4,2.7)} & 1056                      & 176                        & \multicolumn{3}{c}{DMF block(4,3,2)} & 792                       & 132                        & \multicolumn{3}{c}{DMF block(4,2,3)} & 528                       & 88                        \\ \hline
Downsampling                           & \multicolumn{3}{c}{3×3 dwconv}         & -                         & 176                        & \multicolumn{3}{c}{3×3 dwconv}       & -                         & 132                        & \multicolumn{3}{c}{3×3 dwconv}       & -                         & 88                        \\ \hline
5 (7 × 7)                              & \multicolumn{3}{c}{DMF block(4,4,2.7)} & 1056                      & 240                        & \multicolumn{3}{c}{DMF block(4,3,2)} & 792                       & 180                        & \multicolumn{3}{c}{DMF block(4,2,3)} & 528                       & 120                       \\
5 (7 × 7)                              & \multicolumn{3}{c}{DMF block(8,4,2.7)} & 1440                      & 240                        & \multicolumn{3}{c}{DMF block(6,3,2)} & 1080                      & 180                        & \multicolumn{3}{c}{DMF block(4,2,3)} & 720                       & 120                       \\
5 (7 × 7)                              & \multicolumn{3}{c}{DMF block(8,4,2.7)} & 1440                      & 240                        & \multicolumn{3}{c}{DMF block(6,3,2)} & 1080                      & 180                        & \multicolumn{3}{c}{DMF block(4,2,3)} & 720                       & 120                       \\
5 (7 × 7)                              & \multicolumn{3}{c}{DYR - 1×1conv}      & -                         & 1440                       & \multicolumn{3}{c}{DYR - 1×1conv}    & -                         & 1152                       & \multicolumn{3}{c}{DYR - 1×1conv}    & -                         & 960                       \\ \hline
pool \& concat                         & \multicolumn{3}{c}{-}                  & -                         & 1632                       & \multicolumn{3}{c}{-}                & -                         & 1344                       & \multicolumn{3}{c}{-}                & -                         & 1152                      \\ \hline
FC1                                    & \multicolumn{3}{c}{-}                  & -                         & 1920                       & \multicolumn{3}{c}{-}                & -                         & 1920                       & \multicolumn{3}{c}{-}                & -                         & 1600                      \\
FC2                                    & \multicolumn{3}{c}{-}                  & -                         & 1000                       & \multicolumn{3}{c}{-}                & -                         & 1000                       & \multicolumn{3}{c}{-}                & -                         & 1000                      \\ \hline
\# FLOPs                               & \multicolumn{5}{c|}{499M}                                                                       & \multicolumn{5}{c|}{285M}                                                                     & \multicolumn{5}{c}{198M}                                                                     \\ 
\Xhline{3\arrayrulewidth}
\end{tabular}
}
\caption{\textbf{Dynamic Mobile-Former Architectures.} The "bneck-lite" refers to the lite bottleneck block~\cite{li2020micronet}. "DYR - 1$\times$1conv" denotes our dynamic residual point-wise convolution. We employ depth-wise convolution with stride 2 to handle the spatial downsampling. All dynamic residual convolutions used in models use 8 static kernels. "DMF block($H$, $G$, $R$)" denotes Dynamic Mobile-Former block with $H$ heads for the cross attention layer, $G$ groups for the point-wise convolution in DY-Mobile, and $R$ expansion ratio for the IRFFN, respectively.}
\label{tab: architectures}
\end{table*}

%% file: Table/imagenet_results.tex
\begin{table}[]
\centering
\resizebox{1.0\linewidth}{!}{%
\renewcommand{\arraystretch}{1.2}
\begin{tabular}{cccccc}
\Xhline{3\arrayrulewidth}
Model                      & \#Pub    & Reso.   & \#Params & FLOPs & Top-1 \\ \hline
EdgeViT-XXS~\cite{pan2022edgevits}                & ECCV'22  & 256$^2$ & 4.1M     & 557M  & 74.4  \\
EdgeNeXt-XS~\cite{maaz2023edgenext}                & ECCVW'22 & 256$^2$ & 2.3M     & 538M  & 75.0  \\
MobileNetV3 1.0×~\cite{mobilenetv3}           & ICCV'19  & 224$^2$ & 5.4M     & 217M  & 75.2  \\
MobileNetV2 1.0×+ODConv 4×~\cite{li2022omni} & ICLR'22  & 224$^2$ & 11.5M    & 327M  & 75.4  \\  \rowcolor{Gainsboro!60}
\textbf{DMF-XXS}                    & -        & 224$^2$ & 15.1M    & \textbf{198M}  & \textbf{76.0}  \\ \hline
ConT-S~\cite{yan2021cont}                     & arXiv'21 & 224$^2$ & 10.1M    & 1.5G  & 76.5  \\
EfficientNet-B0~\cite{efficientnet}            & ICML'19  & 224$^2$ & 5.3M     & 390M  & 77.1  \\
Swin-1G~\cite{liu2021swin}$^\dagger$                  & ICCV'21  & 224$^2$ & 7.3M     & 1.0G  & 77.3  \\
EfficientViT~\cite{cai2022efficientvit}               & arXiv'22 & 192$^2$ & 7.9M     & 304M  & 77.7  \\ \rowcolor{Gainsboro!60}
\textbf{DMF-XS}                     & -        & 224$^2$ & 19.8M    & \textbf{285M}  & \textbf{77.8}  \\ \hline
PVT-T~\cite{wang2021pyramid}                      & ICCV'21  & 224$^2$ & 13.2M    & 1.9G  & 75.1  \\
EdgeViT-XS~\cite{pan2022edgevits}                 & ECCV'22  & 256$^2$ & 6.7M     & 2.0G  & 77.5  \\
MPViT-T~\cite{lee2022mpvit}                    & CVPR'22  & 224$^2$ & 5.8M     & 1.6G  & 78.2  \\
MobileViT-S~\cite{mehta2021mobilevit}                & ICLR'22  & 256$^2$ & 5.6M     & 2.0G  & 78.4  \\
EfficientNetV2-B0~\cite{tan2021efficientnetv2}          & ICML'21  & 224$^2$ & 7.4M     & 700M  & 78.7  \\
EdgeNeXt-S~\cite{maaz2023edgenext}                 & ECCVW'22 & 224$^2$ & 5.6M     & 965M  & 78.8  \\
BoT-S1-50~\cite{srinivas2021boT}                  & CVPR'21  & 224$^2$ & 20.8M    & 4.3G  & 79.1  \\
CMT-T~\cite{guo2022cmt}                      & CVPR'22  & 160$^2$ & 9.5M     & 600M  & 79.1  \\
Swin-2G~\cite{liu2021swin}$^\dagger$                    & ICCV'21  & 224$^2$ & 12.8M    & 2.0G  & 79.2  \\
MobileFormer-508M~\cite{mobileformer}               & CVPR'22  & 224$^2$ & 14.0M    & 508M  & 79.3  \\ \rowcolor{Gainsboro!60}
\textbf{DMF-S}                      & -        & 224$^2$ & 25.1M    & \textbf{499M}  & \textbf{79.4}  \\ \Xhline{3\arrayrulewidth}
\end{tabular}
}
\caption{ Comparison on ImageNet-1k benchmark. light-weight CNNs, ViT variants, and hybrid models with similar accuracy are grouped together for comparison. The proposed DMFs consistently outperform other models with less computational budget. $^\dagger$ means the results are from ~\cite{mobileformer}.}
\label{tab:imagenet}
\vspace{-1.0em}
\end{table}

%% file: 4_result.tex
\section{Experiments}\label{experiments}
In this section, we evaluate the proposed Dynamic Mobile-Former on ImageNet-1K classification~\cite{deng2009imagenet}, COCO object detection~\cite{lin2014microsoftcoco}, and instance segmentation, by comparing it with representative ViTs, CNNs and their hybrid models.
Ablation analysis is also conducted to showcase the contribution of each novelty in our method.
Our implementation is based on PyTorch library~\cite{paszke2019pytorch} and Timm codebase~\cite{rw2019timm}.

\subsection{ImageNet Classification}
\noindent\textbf{Implementation Details.} ImageNet~\cite{deng2009imagenet} provides approximately 1.28M training and 50K validation images for 1000 categories. 
We train our Dynamic Mobile-Former(DMF) models at an input resolution of 224x224 with a batch size of 1024. 
All models are trained from scratch using AdamW~\cite{loshchilov2017adamw} optimizer for 470 epochs.
We use cosine learning rate schedule~\cite{loshchilov2016sgdr} with linear warmup for 20 epochs.
Data augmentation includes Random Erasing~\cite{zhong2020randomerase}, Horizontal Flip, Random Resized Crop(RRC), and RandAugmentation~\cite{cubuk2020randaugment}.
Further we use multi-scale sampler~\cite{mehta2021mobilevit} during training. 
For regularization, we use weight decay, dropout(different combinations for models with different model configurations), and stochastic depth~\cite{huang2016droppath} with a rate of 0.1.
We also use Exponential Moving average (EMA)~\cite{polyak1992ema} with a momentum of 0.9995 for training.

\vspace{1.5 mm}
\noindent\textbf{Results of DMF.} 
Table~\ref{tab:imagenet} shows the performances of the proposed DMFs that are specified in table~\ref{tab: architectures}.
Our models consistently outperform efficient CNNs and multiple variants of vision transformers, with fewer FLOPs.
In particular, our DMF-XXS achieves a 0.6\% higher top-1 accuracy, while using only 60\% of the FLOPs of MobileNetV2~\cite{mobilenetv2} with ODConv(4$\times$)~\cite{li2022omni}. 
This proves that our dynamic residual convolution with parallel design improves the representational power efficiently.
We also compare our Dynamic Mobile-Former with various ViT variants(EdgeViT~\cite{pan2022edgevits}, Swin~\cite{liu2021swin}, BoT~\cite{srinivas2021boT}, CMT~\cite{guo2022cmt}, MPViT~\cite{lee2022mpvit}).
Specifically, Compared to ~\cite{lee2022mpvit, liu2021swin}, our DMFs achieve higher accuracy but use 3$\sim$4 times less computations.
This is because that DMF uses efficient operators(group convolution~\cite{shufflenet}, dynamic convolution~\cite{chen2020dynamic}) combined with light-weight cross-attention with fewer tokens to extract local features effectively.
Note that our DMF (trained in 470 epochs) even outperforms CMT-Tiny~\cite{guo2022cmt} which leverages much longer training (1000 epochs).
We plot the accuracy-FLOPs curve in Fig.~\ref{fig:plot} (Left) to have an intuitive comparison between these methods.
Prior works on dynamic convolution~\cite{chen2020dynamic, yang2019condconv, li2022omni, decomposition} have shown efficiency, but their performance still falls short compared to low FLOPs regime efficient models.
In contrast, our DMF maximizes representation power combining efficient operators and dynamic convolution in an appropriate  manner, striking a balance between accuracy and FLOPs.

\input{Table/main_ablation}

\input{Table/retinanet}
\input{Table/mask}

\subsection{Object Detection and Instance Segmentation}
\noindent\textbf{Implementation Details.} We validate our DMF as an efficient vision backbone for object detection and instance segmentation with RetinaNet~\cite{lin2017focal} and Mask R-CNN~\cite{he2017mask}, respectively.
The experiments are conducted on COCO~\cite{lin2014microsoftcoco}, which contains 118K training images and 5K validation images of 80 classes.
We pretrain the backbones on the ImageNet-1K and replace the original backbones with our DMFs to generate multi-scale feature maps.
For RetinaNet~\cite{lin2017focal}, All models are trained under standard single-scale and "1$\times$" schedule (12 epochs) from ImageNet pretrained weights.
For Mask R-CNN~\cite{he2017mask}, we train models for both "1$\times$" schedule and "3$\times$" schedule (36 epochs) with a multi-scale training strategy~\cite{liu2021swin}.
We use AdamW~\cite{loshchilov2017adamw} optimizer with an initial learning rate of 0.0001, weight decay of 0.0001, and a batch size of 16. 
We use the popular MMDetection toolbox~\cite{mmdetection} for experiments with all models.

\vspace{1.5 mm}
\noindent \textbf{Results of DMF.}  In Table~\ref{tab: ratina} for object detection with RetinaNet, we compare our Dynamic Mobile-Former with efficient networks (CNNs : ShuffleNetV2~\cite{shufflenetv2}, MobileNetV3~\cite{mobilenetv3}, ResNet~\cite{he2016resnet} ViTs : PVTs~\cite{wang2021pyramid, wang2022pvtv2}, ConTNet~\cite{yan2021cont} Hybrid : MobileFormer~\cite{mobileformer}).
Under similar computational cost, our DMF surpasses ShuffleNetV2~\cite{shufflenetv2}, MobileNetV3~\cite{mobilenetv3} by 9.4 points of mAP and 8.1 points of mAP respectively.
Furthermore, it is worth noting that our DMF consistently outperforms MobileFormer~\cite{mobileformer} by a margin of 0.7 - 1.0\% with less computations. 
For instance segmentation with Mask R-CNN, we report the performance comparison results of instance segmentation in Table~\ref{tab: mask}. Our DMF-S achieves 42.4 AP$^b$, 39.0 AP$^m$, outperforming ResNet50~\cite{he2016resnet} and PVT-Tiny~\cite{wang2021pyramid} which consume more FLOPs (59 - 79G).
This proves effectiveness of the proposed Dynamic Residual Convolution with light-weight attention features.
See Fig. \ref{fig:plot}(Right) for intuitive comparison.

\subsection{Ablations}\label{ablations}
\noindent In this section, we conduct ablation studies on each component of DMF-XXS to investigate the effectiveness of the proposed dynamic residual convolution on ImageNet classification.

\vspace{1.5 mm}
\noindent\textbf{Effectiveness of Dynamic residual convolution.} As shown in Table~\ref{tab:main_ablation}, (a) inputting attention features to kernel attention module and (b) residual connection in kernel space only cost additional 4M FLOPs, but result in a 1.1\% increase in top-1 accuracy over the baseline that use dynamic convolution~\cite{chen2020dynamic}. 
While (a) alone has a relatively small impact on performance, it becomes more effective when used in combination with (b).
It is worth noting that our proposed methods achieve meaningful improvements while incurring only negligible computational costs.

\input{Table/act_ablation}

\vspace{1.5 mm}

\input{Table/init_ablation}

\vspace{1.5 mm}

\input{Table/kernel_num_ablation}

\vspace{1.5 mm}
\noindent\textbf{Activation function for attention scores.} ~\cite{yang2019condconv} proposed the use of sigmoid function to generate attention scores $\pi_k(x)$ in Eq.\ref{eq:our dyconv}. However incorporating sigmoid layer can result in a significantly large kernel space, which can make learning of the attention scores challenging.
~\cite{chen2020dynamic} stabilized the learning process by reducing the kernel space using softmax and demonstrated better performance with fewer kernels and FLOPs.
But, as shown in the table \ref{tab:act_ablation}, we utilize the sigmoid function to enhance performance by increasing the kernel space, as our residual connection in kernel space helps to alleviate optimization difficulties.

\vspace{1.5 mm}
\noindent\textbf{Initialization for $\mathbf{W_{static}}$.} We assume that the previous approach~\cite{chen2020dynamic, yang2019condconv} of using the dynamic convolution module to calculate both input-agnostic and input-dependent kernels in the kernel attention module results in optimization difficulties. Therefore, to address this issue, we introduce the residual connection that connects input-specific dynamic kernel to the existing input-agnostic kernel. The results presented in table \ref{tab:init_ablation} also align with the philosophy of the residual connection~\cite{he2016resnet}.
Note that input-agnostic kernel $\mathbf{W_{input_agnostic}}$ is initialized randomly.

\vspace{1.5 mm}
\noindent\textbf{Static kernel number.} We not only use temperature annealing to achieve stable training, but also utilize residual connection in kernel space to ease the optimization and expand the kernel space. 
Table \ref{tab:kernel_num_ablation} demonstrates that our dynamic residual convolution approach enables us to increase the kernel space by increasing the number of static kernels while maintaining stable training and improved performance.

\noindent Therefore, all variants of DMF take 8 static kernels, which is zero initialized and use sigmoid activation for generating attention scores.

%% file: Table/main_ablation.tex
\begin{table}[t!]
\centering
\resizebox{1.0\linewidth}{!}{%
\tiny
\begin{tabular}{cc|cc|l}
\Xhline{2\arrayrulewidth}
AFKA & RCKS & \#Params & FLOPs  & Top-1 \\ \Xhline{2\arrayrulewidth}
\XSolidBrush   & \XSolidBrush    & 10.0M    & 194.0M & 73.5  \\
\CheckmarkBold    & \XSolidBrush    & 12.6M    & 196.6M & 73.6 (+ 0.1)  \\
\XSolidBrush    & \CheckmarkBold    & 12.4M    & 195.9M & 74.2 (+ 0.7)  \\
\CheckmarkBold    & \CheckmarkBold    & 15.1M    & 198.0M & 74.6 (+ 1.1)  \\ \Xhline{2\arrayrulewidth}
\end{tabular}
}
\caption{ \textbf{Ablation of AFKA and RCKS.} Here, DMF-XXS is used and all models are trained on ImageNet dataset for 250 epochs. "AFKA" and "RCKS" mean Attention Features for Kernal Attention module, Residual Connection in Kernel Space, respectively. }
\label{tab:main_ablation}
\vspace{-1.0em}
\end{table}

%% file: Table/retinanet.tex
\begin{table*}[h!]
\centering
\resizebox{0.75\linewidth}{!}{%
\begin{tabular}{l|cc|ccc|ccc}
\Xhline{3\arrayrulewidth}
Backbone          & \#Params & FLOPs & mAP  & AP$_{50}$ & AP$_{75}$ & AP$_{S}$ & AP$_{M}$ & AP$_{L}$ \\ \hline
ShuffleNet-V2~\cite{shufflenetv2}     & 10.4M    & 161G  & 25.9 & 41.9      & 26.9      & 12.4     & 28.0     & 36.4     \\
MobileNet-V3~\cite{mobilenetv3}      & 12.3M    & 162G  & 27.2 & 43.9      & 28.3      & 13.5     & 30.2     & 37.2     \\
MobileFormer-151M~\cite{mobileformer} & 14.4M    & 161G  & 34.2 & 53.4      & 36.0      & \textbf{19.9}     & 36.8     & 45.3     \\ \rowcolor{Gainsboro!60}
\textbf{DMF-XSS}           & 17.4M    & \textbf{160G}  & \textbf{35.3} & \textbf{54.8}      & \textbf{37.0}      & 19.1     & \textbf{37.9}     & \textbf{47.6}     \\ \hline
ResNet50~\cite{he2016resnet}          & 38.0M    & 239G  & 36.3 & 55.3      & 38.6      & 19.3     & \textbf{40.4}     & 48.8     \\
MobileFormer-294M~\cite{mobileformer} & 16.1M    & 164G  & 36.6 & 56.6      & 38.6      & 21.9     & 39.5     & 47.9     \\
PVT-V2-B0~\cite{wang2022pvtv2}         & 13.0M    & 177G  & 37.2 & \textbf{57.2}      & 39.5      & \textbf{23.1}     & \textbf{40.4}     & 49.7     \\ \rowcolor{Gainsboro!60}
\textbf{DMF-XS}            & 20.2M    & \textbf{161G}  & \textbf{37.3} & \textbf{57.2}      & \textbf{39.8}      & 21.4     & 40.1     & \textbf{50.1}     \\ \hline
ResNet101~\cite{he2016resnet}         & 56.7M    & 315G  & 38.5 & 57.6      & 41.0      & 21.7     & \textbf{42.8}     & 50.4     \\
PVT-V1-Tiny~\cite{wang2021pyramid}          & 23.0M    & 221G  & 36.7 & 56.9      & 38.9      & 22.6     & 38.8     & 50.0     \\
ConTNet-M~\cite{yan2021cont}         & 27.0M    & 217G  & 37.9 & 58.1      & 40.2      & \textbf{23.0}     & 40.6     & 50.4     \\
MobileFormer-508M~\cite{mobileformer} & 17.9M    & 168G  & 38.0 & 58.3      & 40.3      & 22.9     & 41.2     & 49.7     \\ \rowcolor{Gainsboro!60}
\textbf{DMF-S}             & 23.6M    & \textbf{165G}  & \textbf{39.0} & \textbf{59.4}      & \textbf{41.6}      & 22.9     & 42.7     & \textbf{51.4}     \\ \Xhline{3\arrayrulewidth}
\end{tabular}
}
\caption{\textbf{Object detection results on COCO val2017.} All models use RetinaNet~\cite{lin2017focal} as basic framework and are trained on COCO~\cite{lin2014microsoftcoco} train2017 for 12 epochs (1 $\times$) with single-scale training inputs. All backbones are pretrained on ImageNet-1K. The FLOPs(G) are measured at resolution 800$\times$1333.}
\label{tab: ratina}
\end{table*}

%% file: Table/mask.tex
\begin{table*}[h!]
\centering
\resizebox{0.75\linewidth}{!}{%
\begin{tabular}{lcccccccc}
\Xhline{3\arrayrulewidth}
\multicolumn{9}{c}{Mask R-CNN  1×}                                                                                                                                             \\ \hline
\multicolumn{1}{l|}{Backbone}           & \#Params & \multicolumn{1}{c|}{FLOPs} & AP$^b$ & AP$^b_{50}$ & \multicolumn{1}{c|}{AP$^b_{75}$} & AP$^m$ & AP$^m_{50}$ & AP$^m_{75}$ \\ \hline
\multicolumn{1}{l|}{ResNet50~\cite{he2016resnet}}           & 44.0M    & \multicolumn{1}{c|}{260G}  & 38.0   & 58.6        & \multicolumn{1}{c|}{41.4}        & 34.4   & 55.1        & 36.7        \\
\multicolumn{1}{l|}{PVT-V1-Tiny~\cite{wang2021pyramid}}        & 33.0M    & \multicolumn{1}{c|}{240G}  & 36.7   & 59.2       & \multicolumn{1}{c|}{39.3}        & 35.1   & 56.7        & 37.3 
       \\
\multicolumn{1}{l|}{ResNet50 + DyConv~\cite{chen2020dynamic}$^\ddagger$} & 121.8M   & \multicolumn{1}{c|}{260G}  & 39.2   & 60.3        & \multicolumn{1}{c|}{42.5}        & -      & -           & -           \\
\multicolumn{1}{l|}{PVT-V2-B0~\cite{wang2022pvtv2}}          & 23.0M    & \multicolumn{1}{c|}{195G}  & 38.2   & 60.5        & \multicolumn{1}{c|}{40.7}        & 36.2   & 57.8        & 38.6 
       \\ \rowcolor{Gainsboro!60}
\multicolumn{1}{l|}{\textbf{DMF-S}}              & 34.2M    & \multicolumn{1}{c|}{\textbf{181G}}  & \textbf{39.8}   & \textbf{61.6}        & \multicolumn{1}{c|}{\textbf{43.0}}        & \textbf{37.1}   & \textbf{58.8}        & \textbf{39.8}        \\ \hline
\multicolumn{9}{c}{Mask R-CNN  3×}                                                                                                                                             \\ \hline
\multicolumn{1}{l|}{ResNet50~\cite{he2016resnet}}           & 44.0M    & \multicolumn{1}{c|}{260G}  & 41.0   & 61.7        & \multicolumn{1}{c|}{44.9}        & 37.1   & 58.4        & 40.1        \\
\multicolumn{1}{l|}{PVT-V1-Tiny~\cite{wang2021pyramid}}        & 33.0M    & \multicolumn{1}{c|}{240G}  & 39.8   & 62.2        & \multicolumn{1}{c|}{43.0}        & 37.4   & 59.3        & 39.9        \\ \rowcolor{Gainsboro!60}
\multicolumn{1}{l|}{\textbf{DMF-S}}              & 34.2M    & \multicolumn{1}{c|}{\textbf{181G}}  & \textbf{42.4}   & \textbf{63.9}        & \multicolumn{1}{c|}{\textbf{46.3}}        & \textbf{39.0}   & \textbf{61.2}        & \textbf{41.7}        \\ \Xhline{3\arrayrulewidth}
\end{tabular}
}
\caption{\textbf{Instance segmentation results on COCO val2017.} All models use Mask R-CNN~\cite{he2017mask} as basic framework and are trained on COCO~\cite{lin2014microsoftcoco} train2017 for 12 epochs (1 $\times$) with single-scale training inputs and 36 epochs(3 $\times$) with multi-scale training inputs. All backbones are pretrained on ImageNet-1K. The FLOPs(G) are measured at resolution 800$\times$1333. $^\ddagger$ means the results are from ~\cite{li2022omni}.}
\label{tab: mask}
\end{table*}

%% file: Table/act_ablation.tex
\begin{table}[t!]
\centering
\resizebox{1.0\linewidth}{!}{%
\tiny
\begin{tabular}{c|cc|cc}
\Xhline{2\arrayrulewidth}
activation & \#Params & FLOPs  & Top-1 & Top-5 \\ \Xhline{2\arrayrulewidth}
Softmax    & 15.1M    & 198.0M & 72.8  & 90.6  \\
Sigmoid    & 15.1M    & 198.0M & \textbf{73.2}  & \textbf{90.9}  \\ \Xhline{2\arrayrulewidth}
\end{tabular}
}
\caption{ \textbf{Ablation of activation in kernel attention module.} Here, DMF-XXS in used and all models are trained on ImageNet dataset for 150 epochs. }
\label{tab:act_ablation}
\vspace{-1.0em}
\end{table}

%% file: Table/init_ablation.tex
\begin{table}[t!]
\centering
\resizebox{1.0\linewidth}{!}{%
\tiny
\begin{tabular}{c|cc|cc}
\Xhline{2\arrayrulewidth}
Init. method & \#Params & FLOPs  & Top-1 & Top-5 \\ \Xhline{2\arrayrulewidth}
Random       & 15.1M    & 198.0M & 72.9  & 90.8  \\
Zero         & 15.1M    & 198.0M & \textbf{73.2}  & \textbf{90.9}  \\ \Xhline{2\arrayrulewidth}
\end{tabular}
}
\caption{ \textbf{Ablation of initialization method for static kernel.} Here, DMF-XXS in used and all models are trained on ImageNet dataset for 150 epochs. }
\label{tab:init_ablation}
\vspace{-1.0em}
\end{table}

%% file: Table/kernel_num_ablation.tex
\begin{table}[t!]
\centering
\resizebox{1.0\linewidth}{!}{%
\tiny
\begin{tabular}{c|cc|cc}
\Xhline{2\arrayrulewidth}
Kernel number & \#Params & FLOPs  & Top-1 & Top-5 \\ \Xhline{2\arrayrulewidth}
4             & 14.0M    & 196.8M & 72.9  & 90.8  \\
6             & 14.6M    & 197.4M & 73.2  & 90.9  \\
8             & 15.1M    & 198.0M & \textbf{73.6}  & \textbf{91.2}  \\ \Xhline{2\arrayrulewidth}
\end{tabular}
}
\caption{ \textbf{Ablation of static kernel number. } Here, DMF-XXS in used and all models are trained on ImageNet dataset for 150 epochs. }
\label{tab:kernel_num_ablation}
\vspace{-1.0em}
\end{table}

%% file: 5_conclusion.tex
\section{Limitations}
Our Dynamic Mobile-Former (DMF) has fewer FLOPs, but there are other factors that need to be considered for fast inference. For example, if memory access cost is high, then even if FLOPs are low, it may not be fast enough. In terms of memory access cost, the channel expansion in DY-Mobile and IRFFN is a bottleneck for DMF.
For example, first point-wise convolution in DY-Mobile increase channels by 6 times, result in much higher memory access that can cause non-negligible delay and slow down the overall computations. 
Therefore, in future work, we will explore ways to reduce memory while actively utilizing our dynamic residual convolution. Additionally, our convolution module is computationally efficient but not parametrically efficient. Thus, exploring the application of omni-dimensional dynamic convolution~\cite{li2022omni} will also be an important research direction.
Lastly, ~\cite{chen2023fasternet} introduce partial convolution more efficient than group convolution (depth-wise convolution) in terms of FLOPS(short for \textbf{fl}oating-point \textbf{op}erations per \textbf{s}econd).
We will investigate the combination of these newly proposed modules and our methods.

\section{Conclusion} \label{conclusion}
In this paper, we present a Dynamic Mobile-Former(DMF), a model that enhances the potential of dynamic convolution by combining it with efficient operators in a synergistic way.
Our proposed dynamic residual convolution enhances stable learning by adding input-specific kernel to input-agnostic kernel. 
Based on this, we can further increase the model's performance by actively expanding the kernel space.
By integrating lightweight attention and enhanced dynamic convolution, our DMF achieves high efficiency and strong performance on a series of vision tasks.
In particular, DMF demonstrates comparable performance to state-of-the-art models in image classification, and outperforms light-weight CNNs and vision transformer variants with significantly fewer FLOPs in object detection and instance segmentation.
By simplifying the optimization process and effectively utilizing the capabilities of dynamic convolution, our DMF demonstrates the potential for further improvements in mobile vision applications.